\pdfoutput=1

\documentclass[11pt]{article}

\usepackage{acl}

\usepackage{times}
\usepackage{latexsym}

\usepackage[T1]{fontenc}

\usepackage[utf8]{inputenc}

\usepackage{microtype}

\usepackage{verbatim}
\usepackage{multirow}
\usepackage{amsmath}
\usepackage{enumitem}
\usepackage[linesnumbered,ruled,vlined]{algorithm2e}
\SetAlFnt{\small}
\SetAlCapFnt{\small}
\SetAlCapNameFnt{\small}
\usepackage{algorithmic}
\algsetup{linenosize=\tiny}

\usepackage{listings}
\lstdefinelanguage{dict}{
    breaklines=true,
    breakatwhitespace=true,
    basicstyle=\ttfamily\small,
    upquote=true,
    breakindent=0pt,
}

\usepackage{CJKutf8}
\usepackage{graphicx}

\newcommand{\gd}[1]{\textcolor{blue}{}}
\newcommand{\mn}[0]{\textcolor{magenta}{}}
\newcommand{\ac}[1]{\textcolor{purple}{}}
\newcommand{\xn}[1]{\textcolor{red}{}}
\newcommand{\bh}[1]{\textcolor{violet}{}}
\newcommand{\mf}[1]{\textcolor{brown}{}}

\title{CoCoA-MT: A Dataset and Benchmark for Contrastive Controlled MT with Application to Formality}

 \author{Maria N\u{a}dejde, Anna Currey, Benjamin Hsu, Xing Niu, \\ {\bf Marcello Federico}, {\bf Georgiana Dinu} \\
         AWS AI Labs \\ \texttt{mnnadejd@amazon.com}}

\begin{document}
\maketitle
\begin{abstract}
The machine translation (MT) task is typically formulated as that of returning a single translation for an input segment. However, in many cases, multiple different translations are valid and the appropriate translation may depend on the intended target audience, characteristics of the speaker, or even the relationship between speakers. Specific problems arise when dealing with honorifics, particularly translating from English into languages with formality markers. For example, the sentence `Are you sure?' can be translated in German as `Sind Sie sich sicher?' (formal register) or `Bist du dir sicher?' (informal). Using wrong or inconsistent tone may be perceived as inappropriate or jarring for users of certain cultures and demographics.  

This work addresses the problem of learning to control target language attributes, in this case formality, from a small amount of labeled contrastive data. We introduce an annotated dataset (CoCoA-MT) and an associated evaluation metric for training and evaluating formality-controlled MT models for six diverse target languages. 
We show that we can train formality-controlled models by fine-tuning on labeled contrastive data, achieving high accuracy (82\% in-domain and 73\% out-of-domain) while maintaining overall quality.
\end{abstract}

\section{Introduction}
\label{sec:intro}

The quality of neural machine translation (NMT) models has been improving over the years and is approaching that of human translation \cite{DBLP:journals/corr/abs-1803-05567}. 
With fewer glaring accuracy or fluency errors, it is important to address other aspects of translation quality, such as tone and style, in order to generate context-appropriate translations and improve the end-user experience with MT systems. 
In particular, for spoken language and certain text domains (customer service, business, gaming chat), problems arise when translating from English into languages that have multiple formality levels expressed through honorifics or grammatical register. Taking the example from Table~\ref{tab:ContraPairs}, the phrase ‘Could you?’ can have two equally correct German translations: ‘Könnten Sie?’ for the formal register and ‘Könntest du?’ for informal. This problem has been addressed previously with custom models trained on data with consistent formality \citep{ViswanathanWK19}, or through side constraints to control politeness or formality \citep{sennrich-etal-2016-controlling,niu-etal-2018-multi, feely-etal-2019-controlling,schioppa-etal-2021-controlling}. Most prior research has been tailored to individual languages and has labeled large amounts of data using word lists or morphological analysers.  

\begin{table}[t]
\centering
\resizebox{\linewidth}{!}
{

\begin{tabular}{ll}
\hline
 Source & \textit{Could you} provide \textit{your} first name please? \\ \hline
 Informal & \textbf{Könntest du} bitte \textbf{deinen} Vornamen angeben?\\ 
 Formal & \textbf{Könnten Sie} bitte \textbf{Ihren} Vornamen angeben?\\
\hline \hline
Source &  OK, then please \textit{follow} me to your table. \\ \hline
Informal & \begin{CJK}{UTF8}{maru}ではテーブルまで私に\end{CJK}\begin{CJK}{UTF8}{goth}ついて来て。\end{CJK}\\
Formal & \begin{CJK}{UTF8}{maru}ではテーブルまで私に\end{CJK}\begin{CJK}{UTF8}{goth}ついて来てください。\end{CJK}\\
Respectful & \begin{CJK}{UTF8}{maru}ではテーブルまで私に\end{CJK}\begin{CJK}{UTF8}{goth}ついていらしてください。\end{CJK}
\end{tabular}
}
\caption{Contrastive translations for EN-DE and EN-JA with different formality. Phrases in bold were annotated by professional translators as marking formality.\label{tab:ContraPairs}}
\vspace{-2mm}
\end{table}

In this work we look at formality across multiple languages and frame formality control as a transfer learning problem, by leveraging a generic NMT system and a small amount of manually labeled data to obtain MT systems that are controllable for formality. 
Our main contributions are threefold. First, we release a novel multilingual and multi-domain benchmark for \textbf{Co}ntrastive \textbf{Co}ntrolled \textbf{MT} (CoCoA-MT) consisting of contrastive translations with phrase-level annotations of formality and grammatical gender in six diverse language pairs: English (EN) $\rightarrow$ French (FR), German (DE), Hindi (HI), Italian (IT), Japanese (JA), and Spanish (ES). Second, to accompany the CoCoA-MT dataset, we introduce a reference-based automatic metric with high precision at distinguishing formal from informal system hypotheses. Third, we propose training formality-controlled models using transfer learning on contrastive labeled data. Our method is effective across six language pairs and robust across several datasets. We show that transfer learning using CoCoA-MT is complementary to automatically labeled data, while cost-effective compared to non-contrastive curated data.

We release the CoCoA-MT dataset, together with Sockeye 3\footnote{\url{https://github.com/awslabs/sockeye/}} baseline models and evaluation scripts.\footnote{The full data, including train/test splits, will be released at \url{https://github.com/amazon-research/contrastive-controlled-mt/} under a CDLA-Sharing-1.0 license.} These resources were also available to participants of the IWSLT 2022~\citep{iwslt2022} shared task on Formality Control for Spoken Language Translation.

\section{CoCoA-MT Dataset}

We first introduce CoCoA-MT, our \textbf{Co}ntrastive \textbf{Co}ntrolled \textbf{MT} by \textbf{A}WS AI dataset, which enables evaluation and training of formality-controlled models.

\subsection{Source Data}

The EN source data comes from three domains/ modalities: Topical-Chat\footnote{\url{http://github.com/alexa/Topical-Chat/}}~\citep{Gopalakrishnan2019}, as well as new \textit{Telephony} and \textit{Call Center} data.\footnote{The Telephony and Call Center data is part of a larger conversational dataset that is currently a work in progress.} 
Topical-Chat consists of text-based conversations about various topics, such as fashion, books, sports, and music. 
The Telephony domain contains transcribed spoken general conversations, unrestricted for topic. 
The Call Center data is also transcribed spoken data, where the conversations come from simulated customer support scenarios. 

We use these three datasets to extract subsets containing utterances that are relevant to the formality control task. The subsets are designed to ensure coverage of diverse phenomena related to formality (honorifics or grammatical register) in the target languages. 
Specifically, we first selected segments (without the conversational context) having between 7 and 40 words and containing second-person pronouns (relevant for all target languages) and first-person pronouns (relevant for honorifics in JA). Through regular expressions, we ensured that the selected data contained the relevant pronouns in various positions (subject, object, object of preposition). Second, we created a list of common EN verbs and used them in data selection in order to ensure lexical diversity of verbs and verb forms. Third, the automatically selected segments were further filtered or corrected by native English-speaking annotators who were asked to remove stock phrases (e.g.\ \textit{thank you}), ensure that at least one addressee or speaker is referenced, and clean disfluencies from the speech data. 

The selected source segments were then further filtered after the translation and phrase-level annotation steps described in the next section. 



\subsection{Translations and Annotations}
For each source segment, we collected one reference translation for each level of formality (formal and informal). For JA, where more than two formality levels are possible, informal was mapped to \textit{kudaketa}, formal to \textit{teineigo}, and respectful to \textit{sonkeigo} and/or \textit{kenjougo}.\footnote{In this work we only use the informal and formal contrastive translations. For Japanese, we release, when applicable, additional translations for the respectful formality level.} We discarded segments if translators did not provide a translation for each formality level, because we considered these segments not relevant for the formality control task. Table~\ref{tab:ContraPairs} provides examples for EN$\rightarrow$DE and EN$\rightarrow$JA. Annotators also provided phrase-level annotations of formality markers in the target segments in order to facilitate evaluation and analysis (shown in {bold} in Table~\ref{tab:ContraPairs}).

Reference translations were created by professional translators who were native speakers of the specified language and geographic variant.\footnote{For French and Spanish, we release variants from France and Spain, respectively. We will release additional references for Canadian French and Mexican Spanish in the near future.} 
Formal translations were created from scratch as the canonical form, and informal translations were post-edited from the formal translations to ensure that there were no spurious differences between formal and informal references. 
Translators were instructed to generate natural translations that preserve the meaning and tone of the original sentence while addressing formality with minimal required changes. Such changes included swapping pronouns, editing verb forms, and additional lexical changes to obtain natural-sounding translations.
We report dataset statistics in the next section and the full instructions given to translators in Appendix~\ref{sec:app:instructions}. 



\subsection{Dataset Statistics}
For each language pair, we release test data for all three domains (Topical-Chat, Telephony, and Call Center), and training data for Topical-Chat and Telephony. All segments in the test data have distinct formal/informal references, while the training data contains some segments with identical references for both formality levels. 


Table~\ref{tab:DatasetStats} reports the number of training and test segments for each language pair, as well as the overlap (measured as BLEU) between informal and formal references in the test set. Note that EN-JA has more training data because we include both first-person and second-person formality segments. The similarity between formal and informal translations is lowest for EN-JA and highest for EN-HI, confirming that Hindi and Japanese are the two extremes with respect to the degree of formality marking among these six languages. 

\begin{table}[h]
\centering
\small{
\begin{tabular}{lrrr}
\hline
Target & \#train &	\#test & overlap \\
\hline
DE & 400 & 600 & 75.1 \\
ES & 400 & 600 & 79.0 \\
FR & 400 & 600 & 76.7 \\
HI  & 400 & 600 & 81.1 \\
IT & 400 & 600 & 78.8 \\
JA & 1,000 & 600 & 74.6 \\

\hline
\end{tabular}
}
\caption{Number of segments in the training and test data, and overlap between the references in the test set as measured by BLEU (informal vs.\ formal). \label{tab:DatasetStats}}
\end{table}

In Table~\ref{tab:CorpusStats} we report corpus level statistics on the variety of phenomena represented in formal training set, including the number of unique and total phrases and tokens labeled for formality in the reference translation. Additionally, we report on the fraction of tokens labelled for formality that are either verbs or pronomials. To compute the part-of-speech for each token, for Hindi we utilized \texttt{stanza} \cite{stanza-paper}. For the other target languages, we utilized \texttt{spaCy}\footnote{http://spacy.io} and the respective large language models. For Japanese there was a significant number of tokens that were nouns or adjectives (7\%) which was not true for the other target languages (on average 2\%).

\begin{table}[ht]
\centering
\resizebox{\linewidth}{!}
{
\begin{tabular}{l|r|r|r|r|r|r}
\hline
 & \multicolumn{2}{c|}{Phrases} & \multicolumn{4}{c}{Tokens}  \\ \hline
Target & \#unique & \#total & \#unique & \#total & \%VB & \%PR \\
\hline
DE & 183 & 754 & 123 & 1,103 & 35.4 & 64.6    \\
ES & 219 & 625 & 217 & 758 & 48.2 & 42.9    \\
FR & 149 & 624 & 118 & 921 & 35.5 & 60.5    \\
HI & 33 & 627 & 34 & 628 & 18.9 & 80.6 \\
IT & 179 & 615 & 167 & 747 & 43.4 & 52.5    \\
JA & 915 & 2,473 & 619 & 6,778 & 82.6 & 0.0 \\
\hline
\end{tabular}
}
\caption{Formal training set statistics for the phrases and tokens labelled for formality and the fraction of tokens that are verbs or auxillary verbs ("VB") or pronomials ("PR"). Note that for Japanese, roughly 7\% of tokens were either nouns or adjectives.
\label{tab:CorpusStats}}
\end{table}


\section{Formality Evaluation}
In this section, we present a manual analysis of formality expressed in the outputs of two generic commercial systems for inputs sampled from CoCoA-MT. 
Next, we propose and evaluate a reference-based automatic metric which we will later use to evaluate formality-controlled models.

\paragraph{Manual Analysis of Commercial Systems}
General-purpose commercial MT systems are trained on web-scale parallel and monolingual data with different formality levels. To understand how these systems behave with respect to formality, we analyzed two commercial MT systems on 300 random samples from CoCoA-MT. For each target language and each system, two professional translators were asked to label the translations according to the formality markers present in the output: formal, informal, neutral, other. The label ``neutral'' was used for output that can be considered both formal or informal (impersonal-passive or plural forms), while ``other'' was used to label inconsistent formality or incorrectly omitted formality markers\footnote{We give examples in the appendix in Table~\ref{tab:app:manual_eval_labels}.}. 

\begin{table}
\centering
\small{
\begin{tabular}{l l|rrrr|r}

\hline
Lang. & Sys. &	F & I & N & O & IAA \\ \hline
\multirow{2}{1cm}{\centering EN-DE} & A & 45.7 &	46.0& 3.0 & 5.4	& \multirow{2}{*}{0.93} \\
& B &  49.8 & 39.8 & 3.7 & 6.7  &\\ \hline
\multirow{2}{1cm}{\centering EN-ES} & A & 26.8 & 67.4 &	1.5 & 4.2 & \multirow{2}{*}{0.91} \\
& B & 28.0 & 66.1 & 1.2 & 4.7 & \\ \hline
\multirow{2}{1cm}{\centering EN-FR} & A & 68.6 & 24.6 & 0.5 & 6.4 & \multirow{2}{*}{0.94} \\
& B &72.7 & 18.6 & 0.5 & 8.2 & \\ \hline
\multirow{2}{1cm}{\centering EN-HI} & A &	81.7 & 3.2 & 1.7 & 13.5& \multirow{2}{*}{0.96} \\
& B & 87.7 & 5.2 & 1.5 & 5.7 & \\ \hline
\multirow{2}{1cm}{\centering EN-IT} & A &3.7 & 74.9 & 14.4 & 7.0 & \multirow{2}{*}{0.92} \\
& B & 1.3 & 93.3 & 2.7 & 2.7  & \\ \hline
\multirow{2}{1cm}{\centering EN-JA} & A &29.0 & 42.2 & 2.0 & 24.2 & \multirow{2}{*}{0.82} \\
& B & 73.8 & 1.7 & 2.0 & 20.0 &\\ \hline

\end{tabular}
}
\caption{
Percentage of system outputs labeled by professional translators according to the formality level: formal (F), informal (I), neutral (N), other (O).  \label{tab:GenericSystemAnalysis}}
\end{table}

Table~\ref{tab:GenericSystemAnalysis} reports the distribution of labels for the two systems and the inter-annotator agreement measured by Krippendorff’s alpha~\citep{krippendorffalpha}. Agreement is high at 0.91 on average across languages. 
The distribution of formality in the outputs varies widely across languages for both systems. Both systems exhibit cases of inconsistent formality, with over $20\%$ of segments labeled as ``other'' for Japanese. Overall, systems A and B are surprisingly similar in their behaviour, with significant differences in only two languages: system B is more formal than system A for Japanese (73.8\% vs 29.0\%); system A outputs more neutral forms than system B for Italian (14.4\% vs 2.7\%).

 \begin{table}
\centering
\small{
\begin{tabular}{l|cc|cc}
\hline
 &	\multicolumn{2}{c|}{Formal} & \multicolumn{2}{c}{Informal}  \\ \hline	
LP & P & R & P & R  \\ \hline
EN-DE & 0.96 & 0.86 & 0.98 & 0.68  \\
EN-ES & 0.90 & 0.60 & 0.97 & 0.59  \\ EN-FR & 0.98 & 0.78 & 0.94 & 0.66 \\
EN-HI & 0.92 & 0.73 & 0.87 & 0.54 \\
EN-IT & 0.80 & 0.53 & 0.98 & 0.67  \\
EN-JA & 0.71 & 0.43 & 0.69 & 0.54  \\ \hline
\textit{average} & 0.88 & 0.66 & 0.91 & 0.61 \\

\hline
\end{tabular}
}
\caption{Precision and recall of automatic segment-level classification of system outputs as formal or informal. \label{tab:AutomaticMetricEval}}
\end{table} 

\paragraph{Automatic Evaluation}

To evaluate formality-controlled models, we propose a reference-based corpus-level automatic accuracy metric. 
Given a system hypothesis, we automatically label it as formal or informal: formal if the hypothesis contains: a) any of the formality-marking phrases annotated in the formal reference \underline{and} b) none of the phrases annotated in the informal reference. We reverse the conditions to assign an informal label. Note that some hypotheses may not fall into either category.

Following segment-level assignments, we compute a \underline{corpus-level} \textit{Matched-Accuracy} (M-Acc) metric as the percentage of outputs that match the desired formality level, out of all the instances classified automatically as either formal or informal (hence \textit{matched}). We use the notation M-Acc (F)/(I) to denote this score when the desired formality level is formal/informal, respectively. We could not reliably classify neutral and other examples automatically and as such we did not include these labels when computing accuracy. 
Algorithm~\ref{alg:app:MatchedAccuracy} formally describes the implementation of the reference-based automatic \textit{Matched-Accuracy} metric.

\begin{algorithm}[t]
    \SetKwInOut{Input}{Input}
    \SetKwInOut{Output}{Output}
    \Input{System \textit{hypotheses} and annotated (formal, informal) \textit{references}}
    \Output{Matched formal accuracy}
    
    \SetKwData{MatchFormal}{\#match\_formal}
    \SetKwData{MatchInformal}{\#match\_informal}
    \SetKwData{Formality}{formality}
    \SetKwData{NFormal}{\#formal}
    \SetKwData{NInformal}{\#informal}
    \SetKwData{NMatched}{\#matched}
    
    \For{hyp $\in$ hypotheses, (formal\_ref, informal\_ref) $\in$ references}
    {
        \For{marked\_phrase in formal\_ref}
        {
            \If{marked\_phrase in hyp}
            {
                $\MatchFormal\mathrel{+}=1$
            }
        }
        \For{marked\_phrase in informal\_ref}
        {
            \If{marked\_phrase in hyp}
            {
                $\MatchInformal\mathrel{+}=1$
            }
        }

        \uIf{\MatchFormal$>0$ {\bf and} \MatchInformal$=0$}
        {
            $\Formality\leftarrow\texttt{Formal}$
        }
        \ElseIf{\MatchInformal$>0$ {\bf and} \MatchFormal$=0$}
        {
            $\Formality\leftarrow\texttt{Informal}$
        }

        \uIf{\Formality = {\tt Formal}}
        {
            $\NFormal\leftarrow\NFormal+1$
        }
        \ElseIf{\Formality = {\tt Informal}}
        {
            $\NInformal\leftarrow\NInformal+1$
        }
    }
    
    {$\NMatched\leftarrow\NFormal+\NInformal$}
    
    {$\textit{formal\_acc}\leftarrow\NFormal/\NMatched$}
    
    {$\textit{informal\_acc}\leftarrow\NInformal/\NMatched$}
    
    \KwRet{formal\_acc, informal\_acc}
    
    \caption{Algorithm for computing the formal, informal \textit{Matched-Accuracy}. }
\label{alg:app:MatchedAccuracy}
\end{algorithm}

To validate the M-Acc metric, we compare the predictions for formal and informal against the true labels given to outputs of system A and system B (described above).  We report the breakdown of precision and recall for the two labels for each language in Table~\ref{tab:AutomaticMetricEval}. The reference-based segment-level classification algorithm achieves a macro-average of 0.90 precision  and 0.64 recall across formal and informal, with the highest performance for DE (0.97 precision and 0.77 recall) and  the lowest for JA (0.70 precision and 0.49 recall). 

\section{Transfer Learning for Formality Control}
\label{sec:transfer}

We approach formality-controlled NMT as a transfer learning problem, where we fine-tune a generic pre-trained MT model on labeled contrastive translation pairs from the CoCoA-MT dataset. For each source segment we create two labeled training data points: one for each contrastive reference translation (formal and informal). We use a special token with a randomly initialized embedding for the formality label which we attach to the beginning of the source segment.

\begin{figure*}
\begin{center}
\includegraphics[width=\linewidth]{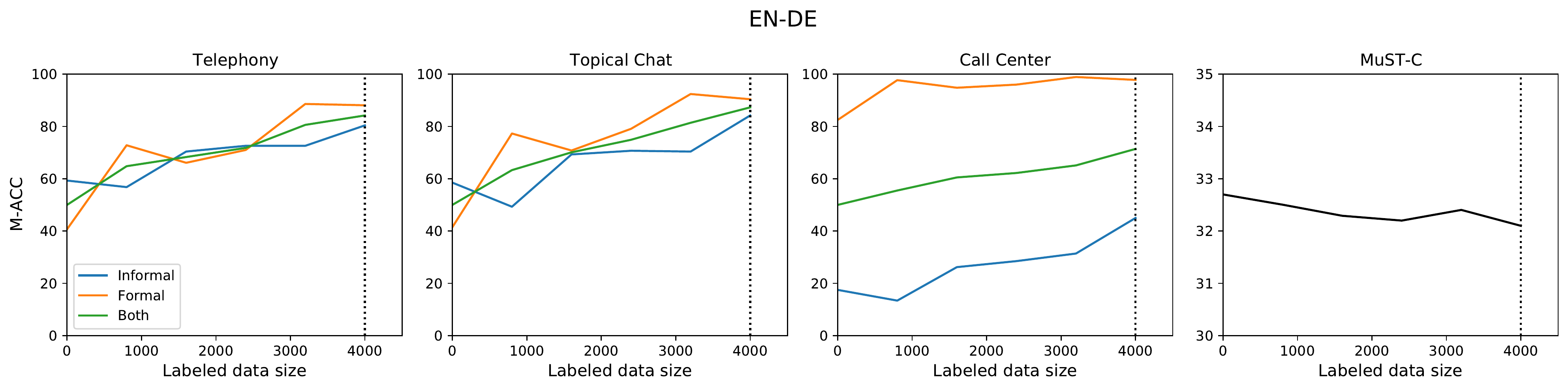}
\end{center}
\caption{Accuracy on the CoCoA-MT test sets and generic quality scores (BLEU) for EN-DE for an increasing amount of the labeled data (up-sampling up to 5x). 
  \label{fig:UpsamplingDataSize}}
\end{figure*}

To leverage the small amount of labeled data while maintaining the overall quality of the generic pre-trained MT model, we first up-sample the labeled data by concatenating multiple copies.\footnote{We study the effect of up-sampling the labeled data in Section~\ref{sec:results}.} Next, we augment the labeled data with an equal amount of unlabeled data sampled randomly from the generic training set. Finally, we fine-tune on the combined labeled and unlabelled data for one epoch with a fixed learning rate following the approach proposed by \citet{hasler-etal-2021-improving} for domain adaptation. With this method we train models that can perform both tasks: generic translation and formality-controlled translation.

\section{Experimental Setup}

NMT models into DE, ES, FR and IT were trained on 20M pairs sampled from ParaCrawl v9 \citep{banon-etal-2020-paracrawl}, using WMT newstest for development. For evaluating generic quality, we used the WMT newstests as well as the MuST-C data \citep{di-gangi-etal-2019-must}.\footnote{We used newstest 2020 for DE, 2014 for ES, 2015 for FR, and 2009 for IT.} 
The EN-JA model was trained on all 10M pairs from JParaCrawl v2 \citep{morishita-etal-2020-jparacrawl} using the IWSLT17 development set. For testing we used WMT newstest2020 and the IWSLT17 test set.
The EN-HI model was trained on all 15M pairs from CCMatrix \citep{schwenk-etal-2021-ccmatrix}, using the WMT newsdev2014 for development and newstest2014 for testing. 

NMT models were built using the Transformer-base architecture \citep{vaswani-etal-2017-attention}, but with 20 encoder layers and 2 decoder layers as recommended by \citet{domhan-etal-2020-sockeye} and SSRU decoder layers for faster decoding \citep{kim-etal-2019-research}.

We report the complete lists of pre-processing and training arguments Appendix~\ref{sec:sockeye_arguments}.


\section{Results}
\label{sec:results}


This section evaluates formality-controlled models trained using CoCoA-MT as described in Section \ref{sec:transfer}. Section \ref{subsec:results1} evaluates performance on CoCoA-MT test sets, investigating in-domain versus out-of-domain performance as well as the effect of up-sampling the labeled data in training. Section \ref{subsec:results2} compares the use of the contrastive CoCoA-MT data with other sources of labeled data. In Section \ref{subsec:results3} we perform additional evaluations on existing (non-contrastive) test sets for which a single formality level is naturally appropriate: forum discussions (informal) and customer support conversations (formal). This is a common scenario, requiring consistent translations that are appropriate for the domain and target audience. 


\subsection{CoCoA-MT Performance}
\label{subsec:results1}

\begin{table*}
\centering
\small{
\begin{tabular}{l  l |ccc|ccc|cc}
\hline

& & \multicolumn{3}{c|}{M-ACC - In-domain} & \multicolumn{3}{c|}{M-ACC - Out-of-domain}  & \multicolumn{2}{c}{BLEU } \\ \hline
Lang. & Labeled data & F & I & Avg. & F & I & Avg. & WMT & TED \\
\hline \hline
\multirow{2}{1cm}{\centering EN-DE} & none & 41.0  & 59.0  & -  & 82.5  & 17.5  & -  & 42.1 & 32.7   \\
 & CoCoA-MT & 89.2  & 82.2  & 85.7  & 97.8  & 45.0  & 71.4  & 41.4 & 32.1   \\ \hline
\multirow{2}{1cm}{\centering EN-ES} & none & 15.9  & 84.1  & -  & 51.5  & 48.5  & -  & 35.1 & 36.7  \\
 & CoCoA-MT & 61.8  & 80.4  & 71.1  & 89.1  & 47.8  & 68.4  & 35.0 & 36.9   \\ \hline
\multirow{2}{1cm}{\centering EN-FR} & none & 89.9  & 10.1  & -  & 100.0  & 0.0  & -  & 38.2 & 43.0   \\ 
 & CoCoA-MT & 76.4  & 61.9  & 69.1  & 98.3  & 13.4  & 55.8  & 39.4 & 45.4   \\ \hline
\multirow{2}{1cm}{\centering EN-IT} & none & 3.6  & 96.4  & -  & 6.4  & 93.6  & -  & 31.5 & 31.4   \\ 
 & CoCoA-MT & 98.5  & 98.2  & 98.4  & 98.5  & 96.5  & 97.5  & 31.7 & 32.0   \\  \hline
\multirow{2}{1cm}{\centering EN-HI} & none & 98.1  & 1.9  & -  & 100.0  & 0.0  & -  & 10.0 &  - \\ 
 & CoCoA-MT & 93.7  & 70.1  & 81.9  & 96.3  & 36.7  & 66.5  & 9.9 & -  \\  \hline
\multirow{2}{1cm}{\centering EN-JA} & none & 64.5  & 35.5  & -  & 65.0  & 35.0  & -  & 21.7 & 14.3  \\
 & CoCoA-MT & 84.8  & 84.4  & 84.6  & 68.8  & 83.2  & 76.0  & 22.2 & 14.3   \\ 
 \hline \hline
 \textit{Average} & CoCoA-MT & 84.1 & 79.5 & 81.8 & 91.4& 53.8 & 72.6& - & - \\ \hline

\end{tabular}
}
\caption{Accuracy of baseline and formality-controlled models on \textit{in-domain} (Telephony, Topical Chat) and \textit{out-of-domain} (Call Center) test splits. The TED test sets are MuST-C for EN-DE,ES,FR,IT and IWSLT for EN-JA. For controlled models, M-Acc (F)/(I) scores are computed using formal/informal translations respectively, resulting in performance upper bounds of 100\%. Baseline un-controlled models generate a single translation, leading to M-Acc (F) and M-Acc (I) to sum up to 100\%.  
\label{tab:AccuracyAllLPs} }
\end{table*} 

To maximize the effectiveness of transfer learning with the small amount of curated labeled data, we first experiment with up-sampling the contrastive labeled data for EN-DE.
Figure~\ref{fig:UpsamplingDataSize} shows accuracy on the CoCoA-MT test sets for different up-sampling factors. We report both formal and informal M-Acc values, obtained by setting the desired formality level to formal/informal and evaluating against formal/informal references respectively. As previously described, the training data covers the Telephony and Topical Chat domains, but not the Call Center domain. For this reason, Telephony and Topical Chat results show in-domain performance while Call Center, out-of domain (distinction also used in Table \ref{tab:AccuracyAllLPs}). BLEU scores are reported as a measure of generic quality: in this setting translations are generated without any formality control.

Results show that by increasing the up-sampling factor (up to 5x), accuracy improves up to 80\% on the combined test sets, while generic quality is fairly stable (small degradation of up to -0.6 BLEU). 
To avoid over-fitting on the labeled data, we fix the up-sampling factor to five for all language pairs throughout the rest of the paper.\footnote{This corresponds to 4,000 total labeled sentence pairs for EN-DE,ES,FR,HI,IT and 10,000 for EN-JA. 
The up-sampling factor can be tuned further for each language to achieve the optimal trade-off between accuracy and generic quality (we report additional results for EN-JA in Appendix~\ref
{sec:app:results}).} When comparing the learning curves for the three domains, we find that Telephony and Topical Chat show similar trends, with high accuracy for both formal and informal, while on Call Center, the out-of-domain setting, the gap between formal and informal accuracy remains large (ca. 50 points). 

Table~\ref{tab:AccuracyAllLPs} reports results on all language pairs. On the in-domain test set, accuracy averaged across formal and informal ranges from  69.1\% for EN-FR to 98.4\% for EN-IT, with generally high accuracy of over 70\% across languages for both formal and informal. 
On the out-of-domain set, accuracy across languages is high for formal (91.4\%) but low for informal (55.4\%). Accuracy for informal is particularly low on this domain for target languages where the generic models have a strong bias toward formal: DE, ES, FR, and HI. 
For these languages, we find this setting adversarial for generating informal outputs as the test set is out-of-domain and at the same time the generic training data biases the models towards formal. We leave for future work exploration of whether increasing data size can overcome this bias, as seems to be the case for EN-JA where informal accuracy is 92.9\%. 
 
From these results we conclude that transfer learning with as little as 400 to 1,000 labeled contrastive examples is effective for formality control on in-domain data and can generalize to out-of-domain data, while generic quality is maintained.\footnote{We observe a side effect on EN-FR where generic quality improves by more than 2 BLEU points on MuST-C. We attribute this to an adaptation effect as both the CoCoA-MT training set and MuST-C test set cover spoken language, which is less represented in web crawled parallel data.} 
 

Finally, a manual investigation of the outputs reveals that formality-controlled models appear to transfer knowledge from the generic training data to generalize to other aspects of formality, beyond grammatical register. We observe examples of changes in lexical choice, punctuation or syntactic structure, even when such variations are not present in the labeled data for that target language. Table \ref{tab:InducedFormalityFeatures} shows some anecdotal examples. We leave a full investigation of this aspect to future work.

\begin{table}[ht]
\centering
{\small{
{
\begin{tabular}{lll}
 \hline
 Src & I am doing well. \textbf{Thanks so much} for asking. \\ \hline
 I & Mir geht es gut. \textbf{Danke so viel}, dass du gefragt hast.\\ 
 F & Mir geht es gut. \textbf{Vielen Dank dafür}, dass Sie fragen.\\
\hline \hline
Src & I will need to know the availability of the day \\ & \textbf{you want to check in}. \\ \hline
I & Tendré que saber la disponibilidad del  día en que \\ & \textbf{quieres hacer el check-in}. \\
F & Tendré que saber la disponibilidad del día en que \\ & \textbf{desea registrarse}. \\
 \hline
\end{tabular}
}
\caption{Examples from formality-controlled models of induced formality features beyond grammatical register. \label{tab:InducedFormalityFeatures}}
}}
\end{table}

\subsection{Effect of Labeled Data Variations}   
\label{subsec:results2}

\begin{table*}
\centering
\small{
\begin{tabular}{l| l l r|ccc|cc}
\hline

 & \multicolumn{3}{c}{Labeled data conditions} &   \multicolumn{3}{c|}{M-ACC - All domains} & \multicolumn{2}{c}{BLEU } \\ \hline
Lang.  & Contrastive & Manual  &\#Src   &F  & I  & Avg.  &WMT  &TED  \\
\hline \hline

\multirow{4}{1cm}{\centering EN-DE}  &  yes & yes   &400   &92.3   &68.6   &80.4   &41.4  &32.1  \\
 & no & yes   &800   &95.7   &76.6   &86.2   &41.2  &32.2  \\
  & no & no &800   &37.3   &77.9   &57.6   &40.8  &31.8  \\
  & no & no & 4,000 & 38.5 & 77.0 & 57.7 & 41.6 & 32.0 \\ \hline
\multirow{4}{1cm}{\centering  EN-ES} & yes & yes  &400    &71.5   &68.8   &70.2   &35.0  &36.9  \\
  & no & yes &800   &65.6   &74.6   &70.1   &35.0  &36.8  \\
 & no & no   &800  &43.4   &78.7   &61.1   &34.9  &36.3  \\ 
 & no & no & 4,000 & 46.1 & 74.3 & 60.2 & 34.7 & 36.2 \\ \hline
\multirow{4}{1cm}{\centering  EN-FR} &  yes & yes  &400   &84.3   &44.2   &64.2   &39.4  &45.4  \\
  & no & yes  &800  &82.5   &47.9   &65.2   &39.1  &45.5  \\
 & no & no  &800   &44.2   &69.3   &56.8   &38.7  &42.6  \\ 
  & no & no & 4,000 & 50.9 & 66.6 & 58.7 & 39.1 & 43.3 \\ \hline
\multirow{3}{1cm}{\centering  EN-JA} & yes & yes  &1,000  & 80.0   & 84.0   &82.0   &22.2  &14.3  \\
  & no & no   &2,000 &44.1   &57.0   &50.6   &21.3  &13.6  \\ 
  & no & no & 10,000 & 45.8 & 54.5 & 50.1 & 21.6 & 13.8 \\ \hline

\end{tabular}
}
\caption{Accuracy of formality-controlled models trained with different sources of labeled data. We consider the fallowing conditions: contrastive vs non-contrastive references and manually vs automatically labeled. The total number of training data points is fixed across all conditions, however the number of unique source segments varies. The TED test sets are MuST-C for EN-DE,ES,FR and IWSLT for EN-JA. 
\label{tab:AccuracyContrastiveConditions}}
\end{table*}

Next, we compare the effectiveness of manually labeled and curated data with that of rule-based automatically labeled data. We create a balanced sample of informal and formal sentence pairs by labeling the target side of the generic training data with methods introduced in prior work: word lists for ES, FR~\citep{ViswanathanWK19}, and JA~\citep{feely-etal-2019-controlling}\footnote{We used the \textit{informal} and \textit{polite} entries from Table 3 of their paper.} and morpho-syntactic rules for DE~\citep{sennrich-etal-2016-controlling}. 
While these methods are known to have limited coverage for languages with complex honorifics systems such as Japanese, or to introduce errors (see examples in Table~\ref{tab:ambiguities}), their advantage is that they can be used to label large amounts of data.\footnote{CoCoA-MT could be used to train a formality classifier that can annotate more data. We leave this to future work.}

\begin{table}[ht]
\centering
{\small{

    \begin{tabular}{ll}
    Source & what are your thoughts on the goatees \\
    & some of the players grow?\\
    Target & ¿qué \textbf{piensas} de las barbas de chivo \\
    & que \underline{se} dejan crecer algunos jugadores?\\
    \end{tabular}
    \caption{Example of an \textbf{informal} sentence from CoCoA-MT classified as \underline{formal} by the rule-based classifier.\label{tab:ambiguities}}
    }}
\end{table}





We compare models trained on the rule-based labeled data against two models: one trained on the contrastive CoCoA-MT data and another trained on \textit{non-contrastive} CoCoA-MT data, with twice as many source segments.\footnote{For EN-JA we did not have additional annotated data for the \textit{non-contrastive} setting.} For comparability, we keep the total number of data points constant across all conditions. However, as additional rule-based labeled data is easy to obtain and may improve results, we test two settings: 800 data points up-sampled 5x (same as the other models), as well as 4000 unique data points. 

Results are shown in Table \ref{tab:AccuracyContrastiveConditions}.
Fine-tuning on noisy rule-based labeled data results in lower average accuracy across all language pairs and significantly worse performance on EN-DE and EN-JA. On FR, DE, and ES, results shift to better informal accuracy with a trade-off in formal performance. For EN-JA the rule-based data is not effective for either formal or informal control. 
Increasing the diversity of the rule-based data by using more unique source segments does not lead to significant improvements. However, given the complementary performance observed, combining the two labeled datasets is a promising future work direction. 

The \textit{non-}contrastive use of CoCoA-MT leads to accuracy improvements of 5.8 points for EN-DE and 1 point for EN-FR. This suggests that improving coverage by sourcing and annotating additional training data is beneficial. However, contrastive data is more efficient to create, as swapping formality levels is done through post-editing. 

\subsection{Human Evaluation on Held-Out Domains}
\label{subsec:results3}

We conduct human evaluation of accuracy and generic quality of formality-controlled models on non-contrastive data from two held-out domains.
The first domain comprises noisy comments on Reddit forums from the MTNT dataset \citep{michel-neubig-2018-mtnt} and the second domain comprises task-based (customer service) dialog turns from the Taskmaster dataset \citep{byrne-etal-2019-taskmaster,farajian-etal-2020-findings}.\footnote{The dialog topics are: ordering pizza, creating auto repair appointments, setting up ride service, ordering movie tickets, ordering coffee drinks and making restaurant reservations. We use the first 35 dialogues included in the WMT 2020 Chat Translation shared task. \url{https://github.com/Unbabel/BConTrasT}} 
For the human evaluation we select source segments that have at least one second person pronoun and set the formality level to informal for the MTNT data and formal for the Taskmaster data, matching the typical formality level used for each domain. Translators were instructed to rate the quality of translations on a scale of 1 (poor) to 6 (perfect) and to mark whether the translation matches the desired formality level. We did not include Hindi as we believed translators would have difficulties with this task given the low level of generic quality (10 BLEU on newstest).

In Table~\ref{tab:human_eval_acc_parity_2}, we report the accuracy and quality scores\footnote{We average the scores of the two annotators and for all sentences.} for the formality-controlled models as well as the improvements over the generic baseline models. Human evaluation results confirm that our formality-controlled models can generalize to unseen domains. Their accuracy is generally high (at or above 70\%) and better than the baseline across languages for both Formal and Informal (with the exception of Formal for French and Japanese). At the same time, generic quality is retained or even slightly improved in some cases (up to 6.9\% for French and 6.4\% for Japanese on Taskmaster) compared to the generic baseline.






\begin{table}
\centering
{\resizebox{\linewidth}{!}{
\begin{tabular}{l|ll|ll
}
\hline
& \multicolumn{2}{c|}{Informal } & \multicolumn{2}{c}{Formal }  \\ 
& Acc$_{bl\_\Delta}$ & Score$_{bl\_\%}$& Acc$_{bl\_\Delta}$ & Score$_{bl\_\%}$\\ \hline
DE& 79.9$_{+30.2}$& 4.3$_{+0.0\%}$ & 90.4$_{+33.3}$& 4.7$_{+0.4\%}$ \\
ES& 75.4$_{+2.0}$& 5.1$_{+0.5\%}$ & 70.1$_{+49.5}$& 5.6$_{+2.2\%}$ \\
FR& 38.0$_{+31.2}$& 4.3$_{+4.0\%}$ & 89.7$_{-3.4}$& 5.0$_{+6.9\%}$ \\
IT& 93.1$_{+21.7}$& 4.3$_{+1.4\%}$ & 92.6$_{+91.9}$& 4.9$_{+2.6\%}$ \\
JA& 80.5$_{+67.5}$& 3.9$_{+3.0\%}$ & 69.1$_{-13.0}$&4.3$_{+6.4\%}$ \\

\end{tabular}
}}

\caption{Human evaluation of formality-controlled models on held-out domains. Formality is set to Informal on MTNT  and to Formal on Taskmaster. We report absolute accuracy difference ($bl\_\Delta$) and relative quality gain ($bl\_\%$) between the controlled and baseline models.}
\label{tab:human_eval_acc_parity_2}
\end{table}




\section{Gender-Specific Translations}
While creating the CoCoA-MT formality-controlled dataset, we observed that for target languages with grammatical gender (all except JA), some reference translations require gender to be expressed in the target even though it is ambiguous in the source.\footnote{Here, we refer to \textit{grammatical} gender of the language; we do not infer or ascribe gender to any speaker or utterance.} 
Table~\ref{tab:GenderEx} shows one such sentence from the EN-ES training set. 

\begin{table}[ht]
\centering\small
{
\begin{tabular}{ll}
\hline
 Source & Did you play with Legos \textbf{growing up}? \\ \hline
 Feminine & ?`De \textbf{peque\~{n}a} jugaba con piezas de Lego?\\ 
 Masculine & ?`De \textbf{peque\~{n}o} jugaba con piezas de Lego?\\
\hline
\end{tabular}
}
\caption{Sentence from CoCoA-MT where grammatical gender is expressed in the target but ambiguous in the source. We show formal translations for illustration.\label{tab:GenderEx}}
\end{table}
\begin{table}[ht]
\small\centering
{
\begin{tabular}{l|rr}
\hline
target & train & test \\\hline
DE & 1\% & 1\%\\
ES & 11\% & 12\%\\
FR & 9\% & 10\%\\
HI & 38\% & 54\%\\
IT & 5\% & 8\%\\
\hline
\end{tabular}
}
\caption{Percent of training and test segments that express gender distinctions in the reference.\label{tab:GenderCts}}
\end{table}

\begin{table*}[ht]
\centering
\small{
\begin{tabular}{l|ccc|ccc|cc}
\hline
 & \multicolumn{3}{c|}{WinoMT} & \multicolumn{3}{c|}{M-ACC - All domains} & \multicolumn{2}{c}{BLEU}\\
model & Acc. & Fem. F1 & Msc. F1 & F & I & AVG & WMT & TED\\\hline
base & 59.5 & 51.2 & {68.0} & 29.0 & 71.0 & - & 35.1 & 36.7\\
msc-trg & 57.7 & 47.9 & 66.9 & {71.5} & {68.8} & {70.2} & 35.0 & 36.9\\
fem-trg & {60.0} & {53.1} & {68.0} & 70.2 & 65.1 & 67.7 & 35.0 & 36.7\\
\hline
\end{tabular}
}
\caption{WinoMT, formality, and BLEU scores on English$\rightarrow$Spanish models trained without formality control (base), and with grammatically masculine and feminine target data.\label{tab:WinomtEs}}
\end{table*}

In fact, this is similar to formality: a grammatical distinction must be made in the target language, even though the source is under-specified with respect to gender. 
Therefore, we create references with feminine and masculine grammatical gender using the same approach as for formality: translators post-edit segments altering only what is necessary to change the grammatical gender.\footnote{We restrict this initial work to two genders because most of the languages examined contain two grammatical genders.} 
This results in up to four translations for each source: \{feminine, masculine\} $\times$ \{formal, informal\}. 
Table~\ref{tab:GenderCts} shows the percent of gendered references in the data for each target language. 


\paragraph{Effect on Gender Translation Accuracy} 
In Section~\ref{sec:results}, for segments with gendered translations, we selected a single gender (in that case, masculine) to use consistently in all training and evaluation data. 
Here, we perform an initial evaluation of the effect of gender-specific formality-controlled data on gender translation accuracy using WinoMT~\citep{stanovsky-etal-2019-evaluating} on EN-ES.\footnote{We evaluate on EN-ES because it is the language pair with the most gendered references of those supported by WinoMT.} 
We compare the baseline (without formality control) to separate models trained using masculine (\textbf{msc-trg}; same as in Table~\ref{tab:AccuracyAllLPs}) and feminine (\textbf{fem-trg}) target data. 
These results, along with formality and quality metrics, are shown in Table~\ref{tab:WinomtEs}. 

Using only masculine target sentences causes a drop in feminine F1, whereas feminine target segments improve feminine F1 without harming masculine F1. 
For easy comparison with Section~\ref{sec:results}, we report formality matched accuracy with respect to the \textit{masculine-reference} test set, which explains the slight drop in formality accuracy for fem-trg. 
These results show that gender-specific translations are useful for maintaining gender translation accuracy when creating formality-controlled models. 


We release the gender-specific translations to open up opportunities to explore the best use of this data. 
The data could also be for gender control given user-specified preferences for gender in translation (similar to formality control explored here). 
We leave these possibilities for future work.

\section{Related Work}
\label{sec:rel}

Controlling politeness for NMT was first tackled by \citet{sennrich-etal-2016-controlling} for EN-DE translation. They appended side constraints to the source text to indicate the preference of verbs or T-V pronoun choices \citep{Brown-and-Gilman-1960} in the output.\footnote{\url{https://en.wikipedia.org/wiki/T-V_distinction}} A similar approach was applied to control the presence of honorific verb forms in EN-JA MT by \citet{feely-etal-2019-controlling}. \citet{ViswanathanWK19} controlled T-V pronoun choices of EN-ES/FR/Czech translations by adapting generic models with T-V distinct data. They collected politeness parallel data using heuristics. 
In a task of FR-EN formality-sensitive MT \citep{niu-etal-2017-study}, translation and EN formality transfer were trained jointly in a multi-task setting \citep{niu-etal-2018-multi,NiuC20}. They assumed cross-lingual formality parallel data is not available and leveraged monolingual formality data instead \citep{rao-tetreault-2018-dear}.

Prior work has also investigated control for attributes besides formality: speaker role and gender \citep{MimaFI97,rabinovich-etal-2017-personalized,ElarabyTKHO18,bentivogli-etal-2020-gender}, voice \citep{yamagishi-etal-2016-controlling}, length/verbosity \citep{takeno-etal-2017-controlling, LakewFWHVBE21}, readability/complexity \citep{StymneTHN13,marchisio-etal-2019-controlling,agrawal-carpuat-2019-controlling}, monotonicity \citep{schioppa-etal-2021-controlling}, translator traits ~\citep{wang-etal-2021-towards} and a writer's proficiency level and native language ~\citep{nadejde-tetreault-2019-personalizing}. Controlling multiple attributes with a single NMT system was investigated by \citet{michel-neubig-2018-extreme,schioppa-etal-2021-controlling}. 
Annotation toolkits or parallel corpora annotated with some of these attributes has also been released, including gender and age \citep{rabinovich-etal-2017-personalized, vanmassenhove-etal-2018-getting, bentivogli-etal-2020-gender}, complexity \citep{agrawal-carpuat-2019-controlling}, and speaker traits \citep{michel-neubig-2018-extreme}. 




\section{Conclusions}

This work addresses the problem of controlling MT output when translating into languages that make formality distinctions through honorifics or grammatical register. To train and evaluate formality-controlled MT models, we introduce CoCoA-MT --a novel multilingual and multidomain benchmark-- and a reference-based automatic metric. Our experiments show that formality-controlled MT models can be trained effectively with transfer learning on labeled contrastive translation pairs from CoCoA-MT, achieving high targeted accuracy and retaining generic translation quality. We release the CoCoA-MT dataset to enable future work on controlling multiple features (formality and grammatical gender) simultaneously. 

\section*{Acknowledgments}
We would like to thank Tanya Badeka and Jen Wang for their valuable help in creating the CoCoA-MT dataset; Weston Feely for helping with the Japanese guidelines; Natawut Monaikul and Tamer Alkhouli for their comments on previous versions of the paper; and the anonymous reviewers for their suggestions.

\section{Ethical Considerations}
\label{impact_statement}

As part of this paper, we created and are releasing formality-controlled contrastive parallel data from English into French, German, Hindi, Italian, Japanese, and Spanish. 
The translations and annotations were created by professional translators who were recruited by a language service provider and were compensated according to industry standards. 
The translations are based on existing English corpora which are not user-generated. 
Before creating the translations, we obtained approval for our use case from the creators of the existing artifacts.

As part of our formality-controlled dataset, we noticed that translations often required the gender of the speaker or the addressee to be specified, even when the English source was gender-neutral. 
As a result, for each such case, we include grammatically feminine and grammatically masculine reference translations. 
We hope that this will open up opportunities for future work in avoiding gender bias when controlling for politeness, and even in improving translations by customizing to the user's desired gender,\footnote{Note that we do not recommend \textit{inferring} gender from the user, but customizing according to user-specified gender.} in a similar way to how we customize for the desired formality in this paper. 
In creating gender-specific reference translations, we limit the differences to words that are grammatically gendered in the target languages, rather than stereotypical or other differences. 
It is important to note that while this paper addresses grammatical gender in translation, it does not use human subjects, infer or predict gender, or otherwise use gender as a variable. 

We would like to emphasize that the work on gender in this paper is very much a work in progress. 
We provide this dataset as an initial contribution; we will continue to improve on this work and this data, and we hope other groups also use and expand on it. 
Most notably, so far we have only produced translations for two genders. 
In the future, we plan on expanding the references translations to more genders, in consultation with native speakers of the target languages and other stakeholders. 
We also would like to analyze gender bias in formality-controlled models, as well as create models that can control for multiple features (e.g., formality and grammatical gender) simultaneously.


\bibliography{anthology,custom}
\bibliographystyle{acl_natbib}

\appendix

\section{Formality Evaluation}
\label{sec:app:automatic_eval}

\paragraph{Manual Analysis}
 We give examples of system outputs labeled as Neutral or Other in Table~\ref{tab:app:manual_eval_labels}.

\begin{table*}
\centering
{\small{
\begin{tabular}{l|l}

 EN & Wow, that's awesome! Who is your favorite Baseball team? I like my Az team lol \\ \hline
 JA & \begin{CJK}{UTF8}{min}うわー、すごいね！おれの好きな野球チームは誰ですか？おれのAZチームは好きです笑 \end{CJK}\\ \hline
 Label & OTHER: \begin{CJK}{UTF8}{min}"すごい" (informal) -- "です" and "好きです" (formal).\end{CJK}\\
 \hline \hline
 EN & You know what I'm saying. You want them to teach you something new. \\ \hline
 DE & Du weißt, was ich meine. Sie möchten, dass sie Ihnen etwas Neues beibringen. \\ \hline
 Label & OTHER: "Du weisst"(informal) -- "Sie möchten" and "Ihnen" (formal) \\ 
 \hline \hline
EN & So I will need an early check-in and if you have a airport shuttle, that will be great. \\ \hline
 IT & Quindi avrò bisogno di un check-in anticipato e se si dispone di una navetta aeroportuale, sarà fantastico. \\ \hline
 Label & NEUTRAL: "si dispone" (impersonal) \\ 

\end{tabular}
}}
\caption{System outputs labeled as "Other" or "Neutral".}
\label{tab:app:manual_eval_labels}
\end{table*}

\section{Additional Results}
\label{sec:app:results}

\begin{figure*}[!h]
\begin{center}

\includegraphics[width=\linewidth]{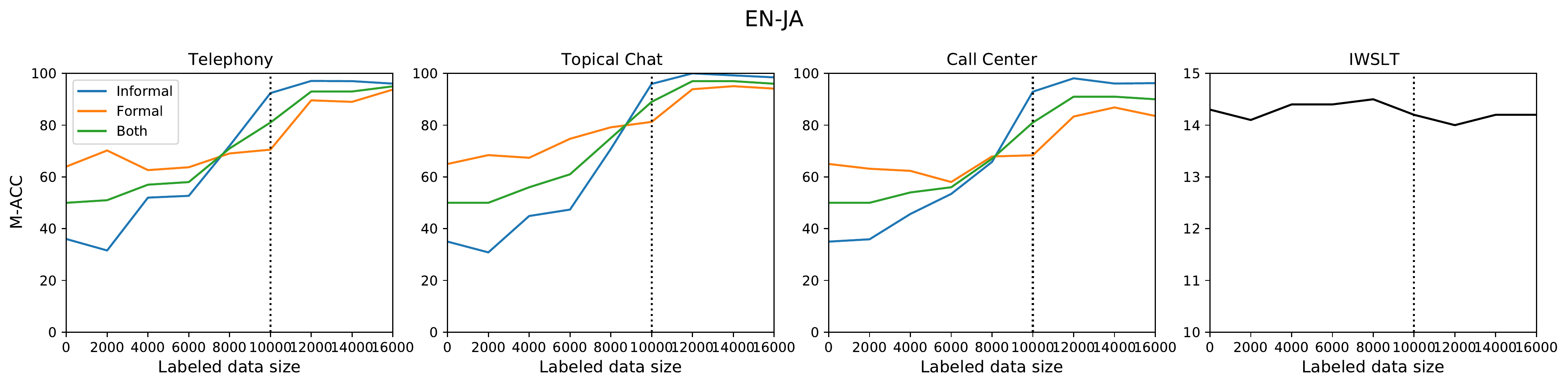}
\end{center}
\caption{Accuracy on the CoCoA-MT test sets and generic quality scores (BLEU) for EN-JA for an increasing amount of the labeled data (through up-sampling up to 8x). The generic baseline scores correspond to 0 on the x-axis. Each source sentence in the CoCoA-MT dataset corresponds to two data points - one for each formality level. For computing the BLEU scores we translate the the generic tes tset(IWSLT) without controling formality.  \label{fig:app:UpsamplingDataSizeJA}}
\end{figure*}

 We report additional results for increasing the up-sampling factor (up to 8x) for EN-JA in Figure~\ref{fig:app:UpsamplingDataSizeJA}. On this larger labeled dataset, a higher up-sampling factor can improve accuracy up to 94\% across domains, significantly increasing the out-of-domain (Call Center) accuracy while generic quality remains stable. The up-sampling factor can be tuned further for each language to achieve the optimal trade-off between accuracy and generic quality.

\clearpage

\section{Experimental Setup}
\label{sec:sockeye_arguments}

All training and development data was tokenized using the Sacremoses tokenizer.\footnote{\url{https://github.com/alvations/sacremoses}} Words were segmented using BPE \citep{sennrich-etal-2016-neural} with 32K operations. Source and target subwords shared the same vocabulary. Training segments longer than 95 tokens were removed.

 The source embeddings, target embeddings, and the output layer's weight matrix are tied \citep{press-wolf-2017-using}. Training is done on 8 GPUs with Sockeye~2's large batch training. It has an effective batch size of 327,680 tokens, a learning rate of 0.00113 with 2000 warmup steps and a reduce rate of 0.9, a checkpoint interval of 125 steps, and learning rate reduction after 8 checkpoints without improvement. After an extended plateau of 60 checkpoints, the 8 checkpoints with the lowest validation perplexity are averaged to produce the final model parameters. 
 
 Fine-tuning is done on 4 GPUs with an effective batch size of 8,192 tokens, a learning rate of 0.0002, and only one epoch, as per \citet{hasler-etal-2021-improving}.

Parameters for standard training:
\begin{lstlisting}[language=dict]
'learning_rate_scheduler_type': 'plateau-reduce', 'keep_last_params': 10, 'update_interval': 16, 'transformer_model_size': (512, 512), 'transformer_postprocess': ('dr', 'dr'), 'learning_rate_warmup': 2000, 'transformer_dropout_act': (0.1, 0.1), 'transformer_feed_forward_num_hidden': (2048, 2048), 'max_num_checkpoint_not_improved': 60, 'weight_init_xavier_factor_type': 'avg', 'optimized_metric': 'perplexity', 'cache_strategy': 'best', 'num_layers': (20, 2), 'use_cpu': False, 'checkpoint_improvement_threshold': 0.001, 'device_ids': [-1], 'learning_rate_reduce_num_not_improved': 8, 'initial_learning_rate': 0.00113, 'seed': 1, 'cache_metric': 'perplexity', 'gradient_clipping_type': 'abs', 'cache_last_best_params': 8, 'weight_init_scale': 3.0, 'dtype': 'float32', 'decode_and_evaluate': 500, 'max_seconds': 1036800, 'amp': True, 'keep_initializations': True, 'transformer_dropout_prepost': (0.1, 0.1), 'transformer_attention_heads': (8, 8), 'weight_tying_type': 'src_trg_softmax', 'learning_rate_reduce_factor': 0.9, 'loss': 'cross-entropy', 'horovod': True, 'num_embed': (512, 512), 'embed_dropout': (0.0, 0.0), 'transformer_preprocess': ('n', 'n'), 'encoder': 'transformer', 'loglevel_secondary_workers': 'ERROR', 'label_smoothing': 0.1, 'batch_size': 2560, 'learning_rate_t_scale': 1.0, 'batch_type': 'max-word', 'optimizer': 'adam', 'transformer_dropout_attention': (0.1, 0.1), 'decoder': 'ssru_transformer', 'min_num_epochs': 1, 'checkpoint_interval': 125, 'transformer_positional_embedding_type': 'fixed', 'lock_dir': '/data', 'gradient_clipping_threshold': -1.0, 'weight_init': 'xavier', 'no_hybridization': False, 'batch_sentences_multiple_of': 8, 'transformer_activation_type': ('relu', 'relu')
\end{lstlisting}

Parameters for fine-tuning:
\begin{lstlisting}[language=dict]
'update_interval': 1,
'learning_rate_warmup': 0,
'checkpoint_improvement_threshold': 0.0,
'initial_learning_rate': 0.0002,
'batch_size': 2048,
'batch_type': 'word'
\end{lstlisting}

\section{Instructions for Creating Formality-Specific References}
\label{sec:app:instructions}

In this section, we reproduce the instructions given to the translators for DE, ES, FR, HI, and IT. 
Instructions for JA are similar but include some language-specific notes. 
We make minor edits for anonymity purposes. 
For brevity, we also remove example translations show to the translators. 

\paragraph{Overview}
This project is to create a test set that content consists of short conversations or utterances taken from conversations. 
Many segments are taken out of context, but all of them are utterances said during a conversation. 
Sometimes you will understand the relationship between the speakers from the context, and sometimes you will not. 

With your translations, we are creating a very specific test set. 
We will use it to test the capability of an MT engine to produce a translation with the \textbf{required} formality of both speakers. 
In other words, imagine if we could ask an MT engine: now translate this conversation as if it is between two speakers, where the relationship between them is formal. 
To test how well it can do that, we will be using your translations (the golden set). 

You will receive a source file that will consist of utterances that were initially part of a conversation; some segments will appear with the surrounding context utterances, and some will be taken out of the conversation. 
Each segment might consist of several sentences. 

\paragraph{Terminology}
\textbf{Formality marker}: a (form of the) word(s) that indicates the tone of that utterance or relationship between speaker and addressee. 
Even if you take this word(s) out of context, by looking at it you will clearly know the tone/formality level of the conversation in which this word is used.

For example, 
\begin{itemize}
    \item \textbf{English}: ``you'' (2nd person pronoun) has no formality marker in English (meaning, by looking at the word you cannot tell if the tone of addressing them is formal or informal). 
    \item \textbf{German}: has formality markers in the 2nd person pronoun and corresponding verb forms - ``du'' (informal) versus ``Sie'' (formal). This means, just by looking at the pronoun ``du'' or the verb next to it, I know the tone is informal. So, I will mark ``du bist'' in DE with Formality tags. 
    \item \textbf{Spanish}: has formality markers for the second person pronouns and their verb conjugations - ``t{\'u}'' (informal) and ``usted'' (formal). Since Spanish is a pro-drop language, verb conjugations may be the only indicator of this information.
    \item \textbf{Italian}: similarly to Spanish, Italian has formality markers for the second person pronouns and their verb conjugations - ``tu'' (informal) and ``lei'' (formal). Since Italian is a pro-drop language, verb conjugations may be the only indicator of this information.
    \item \textbf{French}: similarly to Spanish and Italian, French has formality markers for the second person pronouns and their verb conjugations - ``tu'' (informal) and ``vous'' (formal). French, however, is NOT a pro-drop language.
    \item \textbf{Hindi}: There are Formality markers for 2nd person in Hindi  (meaning, I can address someone respectfully or in a casual way by changing the pronoun). In this case, I will mark the pronoun in Hindi with Formality tags.
\end{itemize}

For Japanese, translators were additionally provided with examples of formality levels (Table~\ref{tab:FormalityLevelsJapanese}) and formality markers (Table~\ref{tab:FormalityMarkersJapanese}).

\begin{table*}
\centering
\small{
\begin{tabular}{l l l l l}
\hline

& Casual speech (\begin{CJK}{UTF8}{maru}常語\end{CJK}jougo)	& \multicolumn{3}{c}{Polite speech (\begin{CJK}{UTF8}{maru}敬語：\end{CJK}keigo)}	\\ \hline

& Jougo / Kudaketa (Informal) & Teineigo (Formal) & Kenjougo (Humble)  & Sonkeigo (Honorific) \\ \hline
Subject for verb & I and you/others	& I and you/others & I & You \\
eat	& \begin{CJK}{UTF8}{maru}食べる \end{CJK}& \begin{CJK}{UTF8}{maru}食べます\end{CJK}& \begin{CJK}{UTF8}{maru}頂く\end{CJK} & \begin{CJK}{UTF8}{maru}召し上がる\end{CJK}\\
come & \begin{CJK}{UTF8}{maru}来る \end{CJK}& \begin{CJK}{UTF8}{maru}来ます\end{CJK} & \begin{CJK}{UTF8}{maru}参上する\end{CJK} &
\begin{CJK}{UTF8}{maru} いらっしゃる \end{CJK}\\
\hline
\end{tabular}
}
\caption{Formality levels in Japanese and examples of changes in inflection or lexical choice for a main verb.\label{tab:FormalityLevelsJapanese}}
\end{table*}

\begin{table*}
\centering
\resizebox{\textwidth}{!}{
\begin{tabular}{l l l l l}
\hline

& Jougo / Kudaketa (Informal) & Teineigo (Formal) & Kenjougo (Humble)  & Sonkeigo (Honorific) \\ \hline

\textbf{1st Person} & & & \\
I'll wait here.	& 
\begin{CJK}{UTF8}{maru}ここで[F]待つ[/F]。\end{CJK}& 
\begin{CJK}{UTF8}{maru}ここで[F]待ちます[/F]。\end{CJK}&
N/A	& 
\begin{CJK}{UTF8}{maru}ここで[F]お待ちします[/F]。\end{CJK}\\
& koko de [F]matsu[/F] & koko de [F]machimasu[/F] & N/A & koko de [F]o-machi shimasu[/F]	\\
\hline

\textbf{2nd Person} & & & \\
What did you buy? &
\begin{CJK}{UTF8}{maru}何を[F]買った[/F]？\end{CJK}&
\begin{CJK}{UTF8}{maru}何を[F]買いました[/F]か？\end{CJK}&
\begin{CJK}{UTF8}{maru}何を[F]お買いになりました[/F]か？\end{CJK}&
N/A	\\
& nani o [F]katta[/F]? & nani o [F]kaimashita[/F] ka? & nani o [F]o-kai ni narimashita[/F] ka? & N/A \\
\hline

\textbf{3rd Person}	& & & \\
The dog chased the cat.	&
\begin{CJK}{UTF8}{maru}犬は猫を[F]追った[/F]。\end{CJK}&
\begin{CJK}{UTF8}{maru}犬は猫を[F]追いました[/F]。\end{CJK}&
N/A	&
N/A	\\
& inu wa neko o [F]otta[/F]	& inu wa neko o [F]oimashita[/F] & N/A & N/A \\
\hline
\end{tabular}
}
\caption{Examples of labeled formality markers in Japanese.\label{tab:FormalityMarkersJapanese}}
\end{table*}

\paragraph{Tags}
We are interested in finding marker words for formality.

\textbf{Formality} tag: \textbf{[F]X[/F]}
NOTE: in Spanish, German, French, Italian and Hindi, Formality is not expressed in the 1st person. 

\paragraph{Task}
There will be two iterations of translating the same source file. Each iteration will be for translating segments into a certain formality. Iteration One will be translated and tagged by Translator 1. Iteration Two will be translated and tagged by Translator 2 (using the Translation Memory from Iteration One).

\paragraph{Steps}
\begin{enumerate}
    \item \textbf{Iteration One - Step One}. Will be done by Translator 1. Translate the segment into the suggested formality. 
    \item \textbf{Iteration One - Step Two}. As you are translating it, think of which words are getting translated with formality markers into your language (if any). \textbf{Tag them in the translation.}
        \begin{enumerate}
            \item \textbf{(!)} You need to first get the translation, then tag the target words that change because of formality - in this order!
            \item It should be the words that have NO formality markers in English - but WILL have formality markers when translated into the target language. These are \underline{ambiguous} source words that only acquire markers in the target language. 
            \item Do not confuse the TONE/STYLE of the utterance overall with the presence of the formality marker words. ``Yo, dang, it’s Sunday already!'' sounds informal overall, but there are no marker words for 1st or 2nd person in Es, De or Hi, so we should NOT tag anything for formality. 
            \item If there are no markers in the translation, do not add any tags, just translate it.
            \item \textbf{NOTE}! If there are ways to translate a sentence with or without markers (for instance using a passive voice), please try to create a natural translation. Do NOT force usage of markers if it creates unnatural translations. 
            \item \textbf{! TIP}: to determine if a word is ambiguous or a marker, take it out of context and see if you can still determine formality.
        \end{enumerate}
    \item \textbf{Iteration 2} (if applicable). This will done by Translator 2. 
        \begin{enumerate}
            \item You will only need to translate and annotate the segments that were tagged with at least one marker during Iteration One. Treat the rest of them as context, where applicable.
            \item One iteration equals one formality level: you will be translating a whole file into one iteration at a time to minimize possible confusion.
            \item Leveraging the Translation Memory from Iteration one, translate the source text into the suggested Formality. 
            \begin{itemize}
                \item For \textbf{Spanish and German}, please try to only change the markers in the translation to the requested Formality combination ( we expect mostly pronouns and verb inflections) and preserve the rest of the translation, if possible. If you disagree with the provided translation or tagging, please raise this to your project manager!
                \item For \textbf{Hindi}, you may have to introduce some additional changes to a sentence when changing its formality levels (choice of words, etc.). Please do that as needed in order to provide a natural translation, but try to be faithful to the source as much as possible.
            \end{itemize}
        \end{enumerate}
\end{enumerate}

Addition for Telephony sourced data: allow the translators to skip the segments that they do not understand/that do not make sense.

\paragraph{Iterations order}
\begin{enumerate}
    \item Formal
    \item Informal
\end{enumerate}

\end{document}